\documentclass[wcp]{jmlr}

\usepackage{longtable}

\usepackage{booktabs}
\usepackage{natbib}

\usepackage{times}
\usepackage{xcolor}
\usepackage{url}
\usepackage{makecell}
\usepackage{graphicx}
\usepackage{pifont}
\usepackage{multirow}
\usepackage{wrapfig}
\usepackage{comment}
\usepackage{amsmath}
\usepackage{makecell}
\usepackage{soul}
\usepackage[utf8]{inputenc}
\usepackage[small]{caption}

\usepackage{footnote}
\makesavenoteenv{tabular}
\makesavenoteenv{table}

\newcommand{\cmark}{\ding{51}}%
\newcommand{\xmark}{\ding{55}}%

\newcolumntype{Y}{>{\centering\let\newline\\\arraybackslash\hspace{0pt}}X}

\jmlrvolume{80}
\jmlryear{2018}
\jmlrworkshop{ACML 2018}

\title{Adversarial TableQA: Attention Supervision  \\
  for Question Answering on Tables}

\author{
\Name{Minseok Cho} \Email{whatjr@yonsei.ac.kr}\\
\addr Yonsei University, Seoul, South Korea
\AND
\Name{Reinald Kim Amplayo}\thanks{Work done while in Yonsei University supported by Yonsei University Research Fund of 2017-22-0082} \Email{reinald.kim@ed.ac.uk}\\
\addr University of Edinburgh, UK
\AND
\Name{Seung-won Hwang}\thanks{corresponding author} \Email{seungwonh@yonsei.ac.kr}\\
\addr Yonsei University, Seoul, South Korea
\AND
\Name{Jonghyuck {Park}} \Email{jhpark0615@ncsoft.com}\\
\addr Knowledge Lab. NCSOFT, South Korea
}

\begin{document}

\maketitle

\begin{abstract}
 The task of answering a question given a text passage has shown great developments on model performance thanks to community efforts in building useful datasets. 
Recently, there have been doubts whether such rapid progress has been based on truly understanding language.
The same question has not been asked in the table question answering (TableQA) task, where we are tasked to answer a query given a table. 
We show that existing efforts, of using ``answers" for both evaluation and supervision for TableQA, show deteriorating performances in adversarial settings
of perturbations that do not affect the answer.
This insight naturally motivates to develop new models  that understand
question and table more precisely.
For this goal, we propose 
\textsc{Neural Operator (NeOp)}, a multi-layer sequential network 
with attention supervision
to answer the query given a table. \textsc{NeOp} uses multiple Selective Recurrent Units (SelRUs) to further help the interpretability of the answers of the model. Experiments show that the use of operand information to train the model significantly improves the performance and interpretability of TableQA models. \textsc{NeOp} outperforms all the previous models by a big margin.
\end{abstract}


\section{Introduction}
\label{sec:intro}

Thanks to the advancements of deep learning for natural language understanding, the task of question answering (QA) has grown since its first occurrence. Specially, QA on structured data has gained more attention from people who want to seek effective methods for accessing these data, as the amount of the structured form of data grows. There are two types of QA tasks on structured data: (1) knowledge base (KB) such as Freebase which is one of the representative the structured data (i.e. KB-QA) \citep{zettlemoyer2005learning,berant2013semantic,cui2017kbqa}, and (2) database such as table which starts to get the spotlight recently (i.e. TableQA) \citep{pasupat2015compositional}. In this paper, we are particularly interested in the TableQA task.

Meanwhile, in the context of TextQA which is the task of answering a question given a text passage, existing systems~\citep{min2017question,liu2017stochastic} have been shown recently,
to be sensitive to adversarially-chosen inputs, such as adding
a distracting sentence that does not affect the answer \citep{jia2017adversarial}.
That is, a new evaluation system, which motivates more robust models that truly understand language, is being considered to be crucial for QA tasks.

\begin{wrapfigure}{r}{0.5\textwidth}
    \centering
    \includegraphics[width=0.5\textwidth]{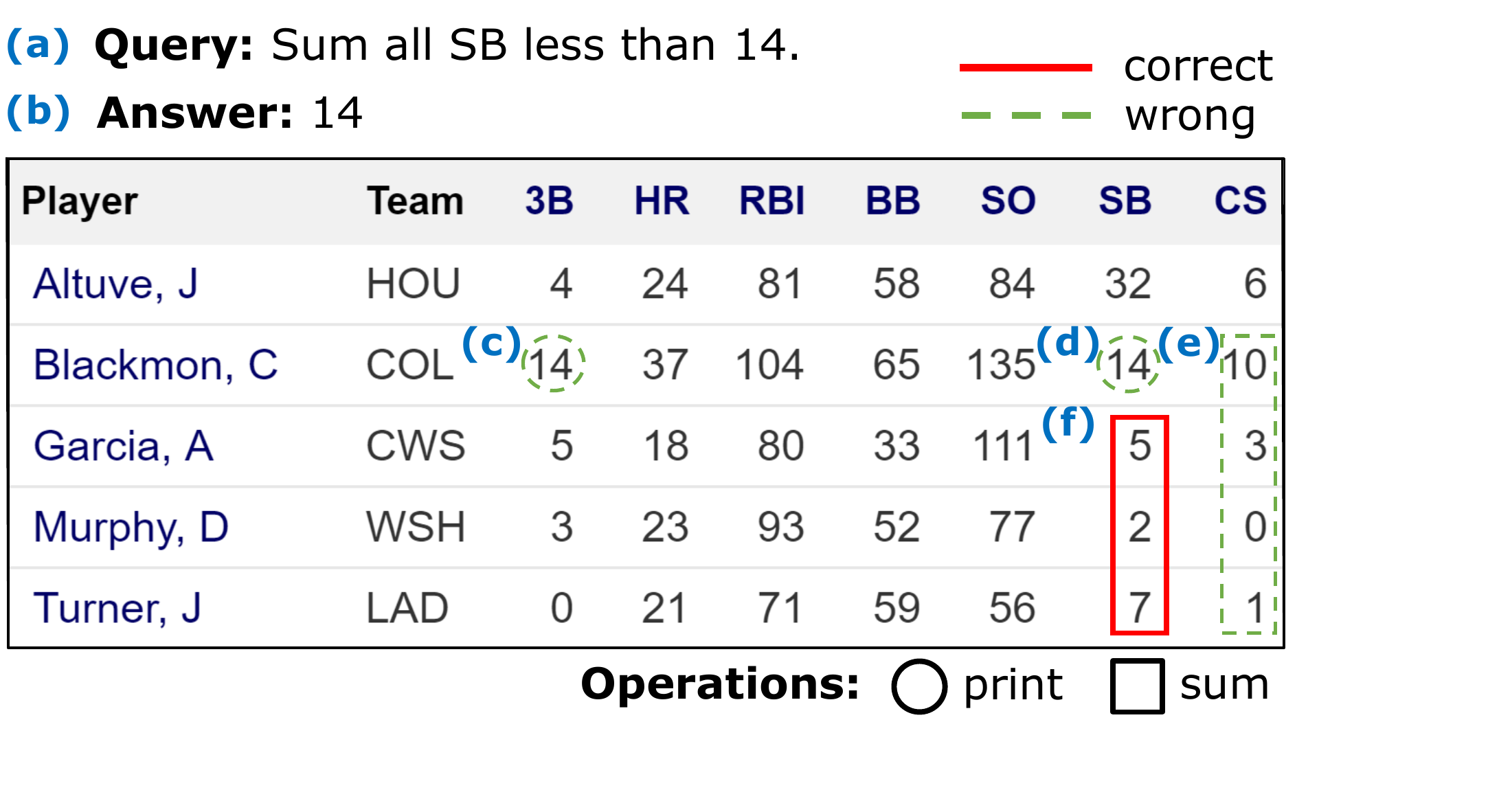}
    \caption{Example data where current TableQA systems may fail.
    Abbreviation glossary: 3B - triple, SB - stolen base, CS - caught stealing.}
    \label{fig:example}
\end{wrapfigure}

The goal of this paper is thus to study an adversarial evaluation system
for TableQA, and QA models that are robust to such attacks.
Current state-of-the-art models in TableQA \citep{pasupat2015compositional,neelakantan2016learning,yin2016neural} are trained upon pairs of natural language questions and answer, 
such as (a) and (b) shown in Figure \ref{fig:example}. 
However, such evaluation may count
``\textit{spurious}'' programs that accidentally lead to the correct answers even by using the wrong cells or operations, as correct answers as also observed in \citep{goldman2017weakly}.
For example, in Figure \ref{fig:example},
the correct program (i.e. (f) in the figure) leads to the correct answer, 
while spurious programs (i.e. (c), (d) and (e) in the figure) also derive the correct answer using wrong reasons.

Based on this observation, we argue that using only answer annotation is too weak for both evaluation and training purposes.
That is, only the query and the final answer are given in the existing datasets, such that no information is given to distinguish the right answer derived from spurious programs.
This leads to confusion in the model on which of the possible ways should it choose to arrive at the final answer. For example, wrong cells can be in the same column (i.e. (d) in the figure), or they can be in a similar column using the same operation (i.e. (e) in the figure).

Recently, SQL statements were proposed as annotations to improve the performance of models \citep{zhongSeq2SQL2017}.
They can uniquely define the answer in Figure~\ref{fig:example}, and hence not spurious, but with the following overheads: 
First, the kind of queries that SQL statements can express is very limited; for example, aggregating values from multiple cells in multiple columns would require a very complicated SQL query that is hard to learn and not supported in the current literature \citep{zhongSeq2SQL2017}.
Second, labeling SQL statements is hard, especially as the text query becomes more complicated, and only few SQL enthusiasts 
would qualify as annotators,
which limits the crowdsourcing of training resources from normal non-technical users.

Because of these reasons, SQL annotations that reflect domain-specific query needs can be obtained, only by a highly  limited set of annotators that are both SQL and domain experts. To relax the former condition, 
the most widely adopted SQL annotated dataset
was created by automatically generating SQL annotations first, such that SQL answers can be translated into corresponding natural language questions by non-technical domain experts (which we call \textbf{annotate-then-query} method). This method dictates domain experts to write questions according to arbitrary generated SQL queries, instead of writing those reflecting their information needs.

In this paper, we propose \textit{attention supervision} that does not require SQL expertise.
Instead, for attention supervision, we propose to use ``operand information'',
the set of correct cells to be selected and operated that provides much richer information about the final answer.
In our running example, aside from feeding the final answer 14 as supervision, we also provide the location of the cells 5, 2, and 7 (i.e. cells of (f) in the Figure \ref{fig:example}).
Operand information is advantageous over both answer and SQL annotations, because of two reasons.
First, all SQL queries (and beyond) can be expressed and annotated using operand information. In the example where multiple cells from multiple columns are aggregated, one can easily mark the cells as operand information.
Second, given the text query, labeling is easy because one only needs to mark the cells to be selected and operated. This makes it easy for any person who understands the query to do annotations directly, guided solely by their curiosity, generating more realistic question workloads than \textbf{annotate-then-query}
datasets, while achieving the scale as effectively.

To this end, our contributions can be divided into three parts.
First, we study the limitation of answer and SQL annotations of existing
datasets and present
\textsc{WikiOps} dataset which annotates operand information.
Second, we present \textsc{Neural Operator} (\textsc{NeOp}), a neural network that learns the correct operands to achieve the final answer. \textsc{NeOp} improves the robustness by  (A) supervising attention weights through the operand information and (B) using three-layer Selective Recurrent Units (SelRUs) to select the following at each timestep: column, pivot, and parameter.
Finally, we compare the performance of \textsc{NeOp} and other baseline models, to show the robustness over adversarial examples.

\section{Related work}

\subsection{Semantic parsing models}

The TableQA task is solvable using semantic parsers which are trained on the question-answer pairs \citep{kwiatkowski2013scaling, krishnamurthy2013jointly}. This method is effective because it does not need expensively annotated pairs of question and program \citep{zettlemoyer2005learning}. However, semantic parsing requires the engineering of hand-crafted and domain-specific grammar or pruning strategies \citep{wang2015building}. The performance of the model is then heavily dependent on the effectiveness of the utilized grammar or pruning strategy \citep{pasupat2015compositional}.

\subsection{Neural models}

Recent methods eliminate the need to use traditional semantic parsing techniques by using neural networks. \textsc{Neural Enquirer} (\textsc{NeEn}) \citep{yin2016neural} is a fully neuralized method to execute a natural language query and return an answer from a table. However, \textsc{NeEn} can only return a single cell as an answer and cannot output an answer from an operation (e.g. sum and average), thus is not able to answer the query in Figure \ref{fig:example}. \textsc{Neural Programmer} (\textsc{NePr}) \citep{neelakantan2016learning} is a neural network model that allows operation on multiple cells, but it only supports three operations: max, min and count.

\subsection{Attention supervision}

Attention supervision has been studied in the areas of visual question answering (VQA) and neural machine translation (NMT) \citep{das2017human,liu2017attention}, where attention that is close to human perception is observed to be more effective for end-task. This observation naturally motivates to explicitly supervise the attention \citep{yu2017supervising,mi2016supervised}  to follow the human's attention.
For QA tasks, the following work can be viewed as a form of attention supervision: \citep{rajpurkar2016squad} introduced spans and \citep{yang2015wikiqa} introduced sentence-type answers for models to easily learn the answer. Recently, \citep{zhongSeq2SQL2017} proposed to use SQL statements as additional supervision knowledgeable at SQL.
In a more general context of NLP tasks, leveraging supervisions beyond the raw text has been effective in text classification \citep{zhao2017topic,Amplayo2018ColdStartAU,Amplayo2018TranslationsAA} and summarization \citep{amplayo2017entity}.

\subsection{Our improvements}

This paper presents \emph{operands} as a unit for both evaluation and training purposes.
First, we build a dataset with  operand annotation, from the manually-generated \textsc{WikiSQL} dataset \citep{zhongSeq2SQL2017} (more discussion in the next section).
Second, our model allows operation on multiple cells and
is extensible to new operations (as we will discuss in Section 4.5).
It also accepts operand information as additional supervision on attention weights.

\section{Our dataset}

\begin{table}[t]
    \footnotesize
    \centering
    {
    \begin{tabular}{|l|l|c|c|c|c|}
        \hline
        Dataset & Size & \makecell[l]{Real Query?} & \makecell[l]{Multi-cell\\Answer?} & \makecell[l]{Real Data?} & \makecell[l]{Attention\\Supervision?} \\ \hline
        \makecell[l]{\textsc{WikiTableQuestions}*\\ \citep{pasupat2015compositional}} & \makecell[l]{22K queries\\2K tables} & \cmark & \cmark & \cmark & \xmark \\ \hline
        \makecell[l]{\textsc{NePr} data\\ \citep{neelakantan2015neural}} & Unlimited & \xmark & \cmark & \xmark & \xmark \\ \hline
        \makecell[l]{\textsc{NeEn} data\\ \citep{yin2016neural}} & Unlimited & \xmark & \xmark & \xmark & \xmark \\ \hline
        \makecell[l]{\textsc{WikiSQL}\\ \citep{zhongSeq2SQL2017}} & \makecell[l]{81K SQL queries\\24K tables} & \xmark & \cmark & \cmark & SQL Statements \\ \hline
		\textbf{\textsc{WikiOps}* (ours)} &  \textbf{\makecell[l]{81K operands\\24K tables}} & \xmark & \textbf{\cmark} & \textbf{\cmark} & \textbf{\makecell[l]{Operand\\ Information}} \\ \hline
        \textbf{\textsc{MLB}* (ours)} & \textbf{\makecell[l]{36K operands\\2K tables}} & \xmark & \textbf{\cmark} & \textbf{\cmark} & \textbf{\makecell[l]{Operand\\ Information}} \\ \hline
    \end{tabular}
    }
    \caption{A survey and comparison of previous TableQA datasets and our proposed  \textsc{WikiOps} dataset. Datasets with asterisk (*) are publicly shared.
    }
    
    \label{tab:datacomp}
\end{table}

We survey recent TableQA datasets and compare the characteristics of the data used in Table \ref{tab:datacomp}.
Current public datasets can be categorized into real and synthetic queries.
Only dataset that collects real queries from users is
WikiTableQuestions \citep{pasupat2015compositional}, which has the advantage of reflecting real query needs.
However, such manual annotation limits the query size small in size,
thus neural models cannot be trained.
On the other hand, automatically generated queries can be arbitrarily large, but cannot reflect real-world distributions of queries. Some datasets generate data tables synthetically as well, enjoying the unlimited size of query and data size, with the expense of them being less realistic.
Recently, a dataset called \textsc{WikiSQL} \citep{zhongSeq2SQL2017}, which contains real-life tables, but not realistic questions, because, as discussed earlier, automatically generated SQL queries are annotated first, dictating to annotate matching questions  (or, \textbf{annotate-then-query}).

We perform preliminary analysis based on the current datasets, with the following insight on the gap of real-life and synthetic queries.

\begin{enumerate}
    \item More than half of the real queries in WikiTableQuestions \citep{pasupat2015compositional} are only solvable using multiple cells, to sum all home runs of players with a certain position. All previous models assuming a single cell assumption, such as \textsc{NeEn}, thus cannot support real-life query needs.

    \item WikiTableQuestions queries may include unconventional aggregations, such as getting the ranges, top 3 values, or difference.
    Desirable models should rapidly adapt to domain-specific operation, unstudied in all existing models.
    
    \item  Although providing additional supervision helps, SQL statements provide limitation on the kind of expressions used as query. As an example, the query ``sum the total second bases, third bases, and homeruns of player A'' would need to calculate the sum cells from multiple columns. This is difficult to express as an SQL statement.
    Apart from that, labeling instances with SQL statements is very hard for non-technical users.

\end{enumerate}

\begin{table*}[t]
  \footnotesize
  \centering
    \begin{tabular}{|cccm{12em}|r|}
    \hline
    \multicolumn{1}{|c|}{Type} &
      \multicolumn{1}{c|}{Operation} &
      \multicolumn{1}{c|}{Query example} &
      \multicolumn{1}{c|}{Logical form} &
      \multicolumn{1}{c|}{\#Query}
      \\
    \hline
    \multicolumn{1}{|c|}{\multirow{1}[7]{*}{Print}} &
      \multicolumn{1}{c|}{\multirow{1}[7]{*}{\texttt{all}}} &
      \multicolumn{1}{p{17em}|}{- What positions does the college/junior/club team, molot perm have?} &
      \multirow{1}[7]{*}{\makecell[l]{\texttt{all(Pos|Team=}\\ {} {} \texttt{Molot Perm)}}} & 
      \multirow{1}[7]{*}{58K}
      
      \\
    \hline
    \multicolumn{1}{|c|}{\multirow{2}[11]{*}{Extremity}} &
      \multicolumn{1}{c|}{\multirow{1}[7]{*}{\texttt{min}}} &
      \multicolumn{1}{p{17em}|}{- What's the 2001 census of the region of Abruzzo where the 1981 census is bigger than 51092.0?} &
      \multirow{1}[3]{*}{\makecell[l]{\texttt{min(2001 Census|Reg=}\\ {} {} \texttt{Abruzzo,1981 Census}\\ {} {} \texttt{>51092.0)}}} & 
      \multirow{1}[7]{*}{5K}
      
      \\
    \multicolumn{1}{|c|}{} &
      \multicolumn{1}{c|}{\multirow{1}[3]{*}{\texttt{max}}} &
      \multicolumn{1}{p{17em}|}{- The record of 7-3 had the largest attendance of what?} &
      \multirow{1}[0]{*}{\makecell[l]{\texttt{max(Attendance}\\ {} {} \texttt{|Record=7-3)}}} & 
      \multirow{1}[3]{*}{5K}
      
      \\
    \hline
    \multicolumn{1}{|c|}{\multirow{2}[11]{*}{Totality}} &
      \multicolumn{1}{c|}{\multirow{1}[7]{*}{\texttt{count}}} &
      \multicolumn{1}{p{17em}|}{- What is the total number of music genre/style in which the lyrics are a detective?} &
      \multirow{1}[7]{*}{\makecell[l]{\texttt{count(Genre|Theme}\\ {} {} \texttt{=Detective Story)}}} & 
      \multirow{1}[7]{*}{7K}
      
      \\
    \multicolumn{1}{|c|}{} &
      \multicolumn{1}{c|}{\multirow{1}[3]{*}{\texttt{sum}}} &
      \multicolumn{1}{p{17em}|}{- What is the total number of points team Alfa Romeo 184T won?} &
      \multirow{1}[0]{*}{\makecell[l]{\texttt{sum(Points|Team=}\\ {} {} \texttt{Alfa Romeo 184T)}}} & 
      \multirow{1}[3]{*}{3K}
      
      \\
    \hline
    \multicolumn{1}{|c|}{\multirow{1}[3]{*}{Centrality}} &
      \multicolumn{1}{c|}{\multirow{1}[3]{*}{\texttt{avg}}} &
      \multicolumn{1}{p{17em}|}{- Tell me the average attendance for week of 11} &
      \multirow{1}[3]{*}{\texttt{avg(Attend|Week=11)}} & 
      \multirow{1}[3]{*}{3K}

      \\
    \hline
    \multicolumn{4}{|c|}{Total Queries} & 81K
      
      \\
    \hline
    \end{tabular}%
  \caption{\textsc{WikiOps} dataset statistics per operation grouped into types with corresponding sample queries}
  \label{tab:wikisql_portion}%
\end{table*}

To compare and contrast SQL with its  operand annotations, we present \textsc{WikiOps}\footnote{We share \textsc{WikiOps} dataset here: \url{https://github.com/MinseokCho/NeuralOperator}}, an altered version of the original \textsc{WikiSQL} dataset \citep{zhongSeq2SQL2017}, by transforming the SQL statements into operand information. 
This can be done as follows. For the print operation, we execute the original SQL statements. For other operations, we modify the SELECT statement by removing the aggregation function (e.g. SELECT SUM(a) to SELECT a).
Sample data instances are shown in Table \ref{tab:wikisql_portion}. We note that this dataset is skewed towards having many \texttt{all} operations and the total number of operations are smaller.

Meanwhile, to motivate against the practice of \textbf{annotate-then-query} of \textsc{WikiOps}, we also crowdsourced 21 natural language question templates from baseball experts 
and annotate operands
to answer such queries, 
over subsets of player stats tables gathered from the MLB website.
Collected expert queries include
complex patterns (e.g. ``give me the smallest'', ``what is the minimum'', etc.).
We call this dataset, \textsc{MLB} dataset, paraphrased into  36k questions.
We note that this dataset, being generated from expert questions,
is expected to  reflect more realistic question patterns of the given domain.

\section{Our model: \textsc{Neural Operator}}

In this section, we present our model \textsc{Neural Operator} (\textsc{NeOp}). \textsc{NeOp} is a multi-layer sequential network that accepts as input the query and the given table, and learns both the operands and the operation to calculate the final answer. In Figure \ref{fig:model_ex}, we take an example to explain how to answer the query in our model. When the table is given, any query can be divided to a sequence as follows:

\begin{wrapfigure}{r}{0.5\textwidth}
    \centering
    \includegraphics[width=0.5\textwidth]{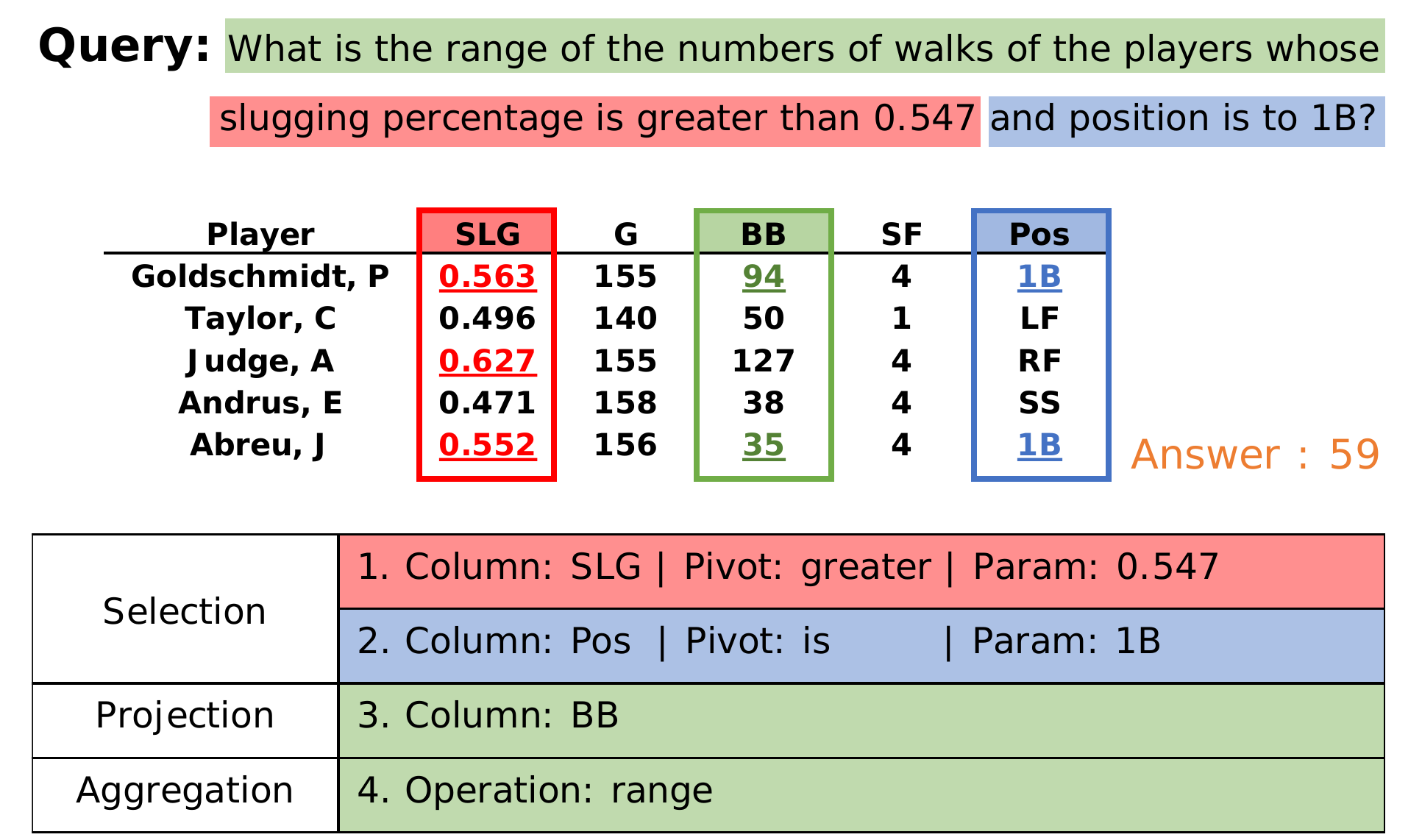}
    \caption{An example query ran through \textsc{NeOp}.
    'number of walks', 'slugging percentage' and 'position'  corresponds to the column 'BB', 'SLG' and 'Pos' in the baseball, respectively.}
    \label{fig:model_ex}
\end{wrapfigure}

\begin{enumerate}
    \item Selection: select a column in the table and words in the query as pivot and parameter\footnote{Here on, we define \textit{pivot} as the conditional operator (e.g. greater) and \textit{param} as the conditional operand (e.g. $0.547$).}, then output the corresponding rows. (Section \ref{sec:selru} and \ref{sec:rowrnn})
    \item Projection: project the selected column to the selected rows, then output the qualifying cells as operand information. (Section \ref{sec:opsel})
    \item Aggregation: execute an aggregate operation through the projected cells from step (2) above. (Section \ref{sec:opsolve})
\end{enumerate}

As shown in the Figure \ref{fig:model_ex}, our model can interpret the answer through showing the highlighted words and column at each sequence. This process is very intuitive for users who want to know how the answer is generated.
The full architecture of the model is shown in Figure \ref{fig:neop}.

\subsection{Encoding}

\paragraph{Word and cell encoding}

We use a single word embedding matrix $W_e = \{w\}$ for the initial vectors of both space-separated strings in the query and cells in the table. A word vector $w_i$ is initially encoded using the concatenation of two encoding methods: (1) an embedding $w_{i_x}$ initialized using Xavier initialization \citep{glorot2010understanding}, and (2) a binary encoding $w_{i_b}$ of the equivalent numerical value of the word, i.e. $w_i = [w_{i_x}; w_{i_b}]$. If the word does not have an equivalent numerical value (i.e. a plain string), then the binary encoding is a zero vector. If the word is a numerical value in the form $int.dec$ where $int$ is the integer part and $dec$ is the decimal part of the number, the binary encoding $w_{i_b}$ is a concatenation of the binarized value of the integer part and the decimal part, i.e. $w_{i_b} = [int_2; dec_2]$. In experiments, we use 300 dimensions each for the word encoding and binary encoding and 15 dimensions for decimal part of the number. Our intention behind choosing 15 is to represent most numbers using 16 bits for decimal part.
Our initial experiments showed that the inclusion of the binary encodings increase the performance of the model.

\paragraph{Query encoder}

A query $q$ is a natural language text composed of multiple words $q_1, q_2, ..., q_l$ where $l$ is the length of the query, and these words are represented by $w_{q_1}, w_{q_2}, ..., w_{q_l}$ from word embedding matrix $W_e$. As used in practice \citep{chung2014empirical}, we encode the query using bi-directional GRU \citep{bahdanau2014neural}, i.e. $\overrightarrow{h}_i, \overleftarrow{h}_i = GRU(w_{q_i}, \overrightarrow{h}_{i-1}), GRU(w_{q_i}, \overleftarrow{h}_{i+1})$. The forward and backward final states are concatenated to create the query vector, i.e. $q = [\overrightarrow{h}_l; \overleftarrow{h}_1]$. 

\paragraph{Table encoder}

A table $T$ consists of $n$ rows and $m$ columns and a total of $n \times m$. $T[j,k]$ is a cell value belonging to the $j$th row and the $k$th column and it is initialized as $w_{T[j,k]}$ from the word embedding matrix $W_e$. Moreover, each column header is represented by a field embedding $f_k$ representing each column.
The field embedding $f_k$ is also obtained from the word embedding matrix $W_e$.
Finally, we follow the method of updating the cell vectors $c_{j,k}$ of \citep{yin2016neural}, where $w_{T[j,k]}$ is updated to include the information about the field it is currently in, i.e. $c_{j,k} = tanh(W_t [w_{T[j,k]}; f_k] + b)$.

\subsection{Selective Recurrent Units}
\label{sec:selru}

Previous neural-based TableQA models used coarse-granular representations
for cells to operate, which
lessens the interpretative power of the model. For example, \textsc{Neural Enquirer} \citep{yin2016neural} only used a single column vector, thus it is hard to know which condition and operation are selected at each timestep.
\textsc{Neural Programmer} \citep{neelakantan2015neural} improves this by using separate vectors,
however these vectors are \textit{simultaneously} created. 

\begin{figure}[t]
    \centering
    \includegraphics[width=1.0\textwidth]{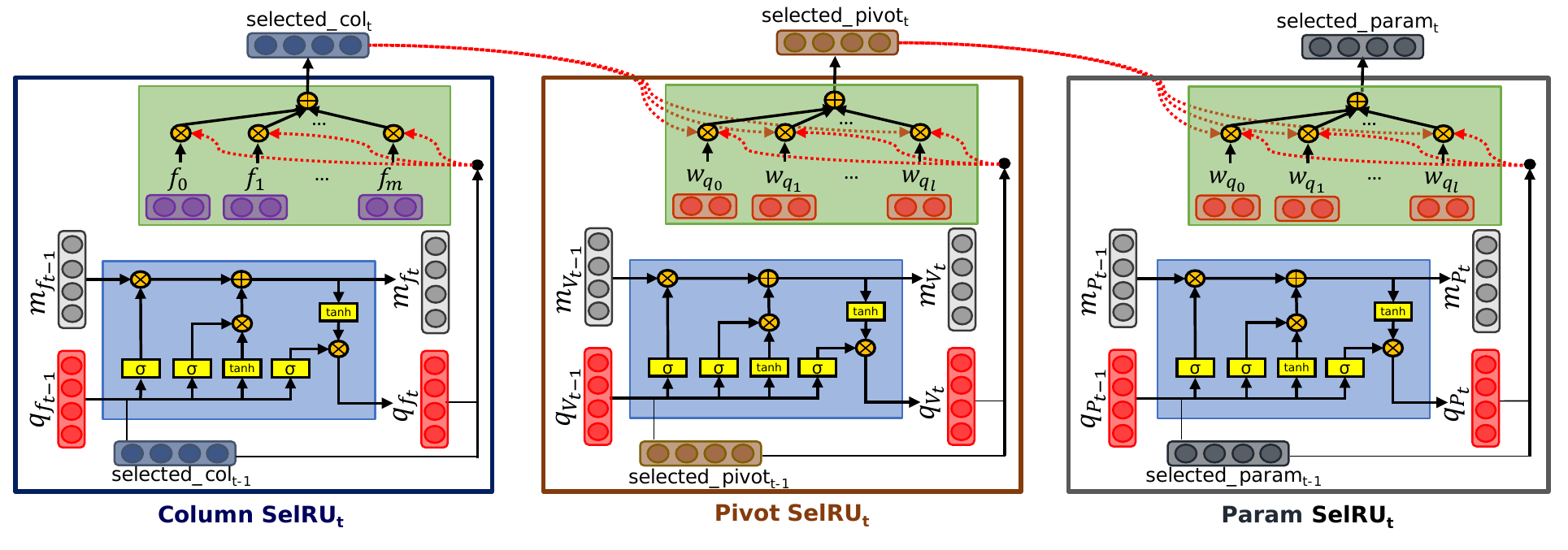}
    \caption{At timestep t, Column, Pivot, and Param SelRUs and their sub-components. All SelRUs have both LSTM and the Attentive Pooling sub-components. Pivot and Param SelRU have one more parameter than Column SelRU, received from the previous SelRU layer.}
    \label{fig:selru}
\end{figure}

We argue that the selection of three different elements in a condition, i.e. columns, pivots and parameters, should be \textit{cascaded}.
For example, in Figure \ref{fig:model_ex}, we first select the column \texttt{SLG}. Then, we use the column information \texttt{SLG} as context to select the pivot \texttt{greater}. Finally, we use the pivot information \texttt{greater} to select the parameter \texttt{0.547}. 
These vectors are leveraged to select the correct rows at each timestep.
Since the model explicitly separates the selection of the parts of the condition, interpreting the model becomes clearer, as we show in Section \ref{sec:interpret}.

We propose Selective Recurrent Units (SelRUs), shown in Figure \ref{fig:selru}. SelRU is composed of two sub-components: (1) an LSTM layer for processing the query vector to contain the semantic information of the query as well as the previous selection information, 
and (2) an attentive pooling component to focus on the current information. We use three SelRUs in turn: Column SelRU, Pivot SelRU, and Param SelRU to select the column, pivot, and parameter, respectively, at each timestep.
In example of Figure \ref{fig:model_ex}, for column, previous selection information is \texttt{SLG} and \texttt{Pos} at timestep $3$.

\textbf{Column, Pivot and Param SelRUs} output the selected column vector $\widetilde{f}_{t}$, selected pivot vector $\widetilde{v}_{t}$ and selected param vector $\widetilde{p}_{t}$ respectively for selecting each element at timestep $t$. To get each selected instance vector, 1) LSTM layer encodes query vector to contain the previous selection information, 2) attentive pooling component yields selected instance vector from newly encoded query vector and other inputs.

First, we can compute latent representations of query and memory:
\begin{equation}
  \label{eq:lstm}
  q_{f_{t}}, m_{f_{t}} = LSTM(\widetilde{f}_{t-1}, q_{f_{t-1}}, m_{f_{t-1}})
\end{equation}
Here $q_{f_t}$ is query vector carrying semantic information of the query and previous selection information of column, and $m_{f_t}$ is memory vector which save previous selection information. 
Because we have newly encoded the query to contain the previous selection information, we can use a different query vector for each timestep. This helps to select the correct condition and row. 
At timestep $0$, $q_{f_0}$ is GRU-encoded vector $q$ and $m_{f_0}$ is initialized to zero vector.

\begin{figure*}[t]
    \centering
    \includegraphics[width=\textwidth]{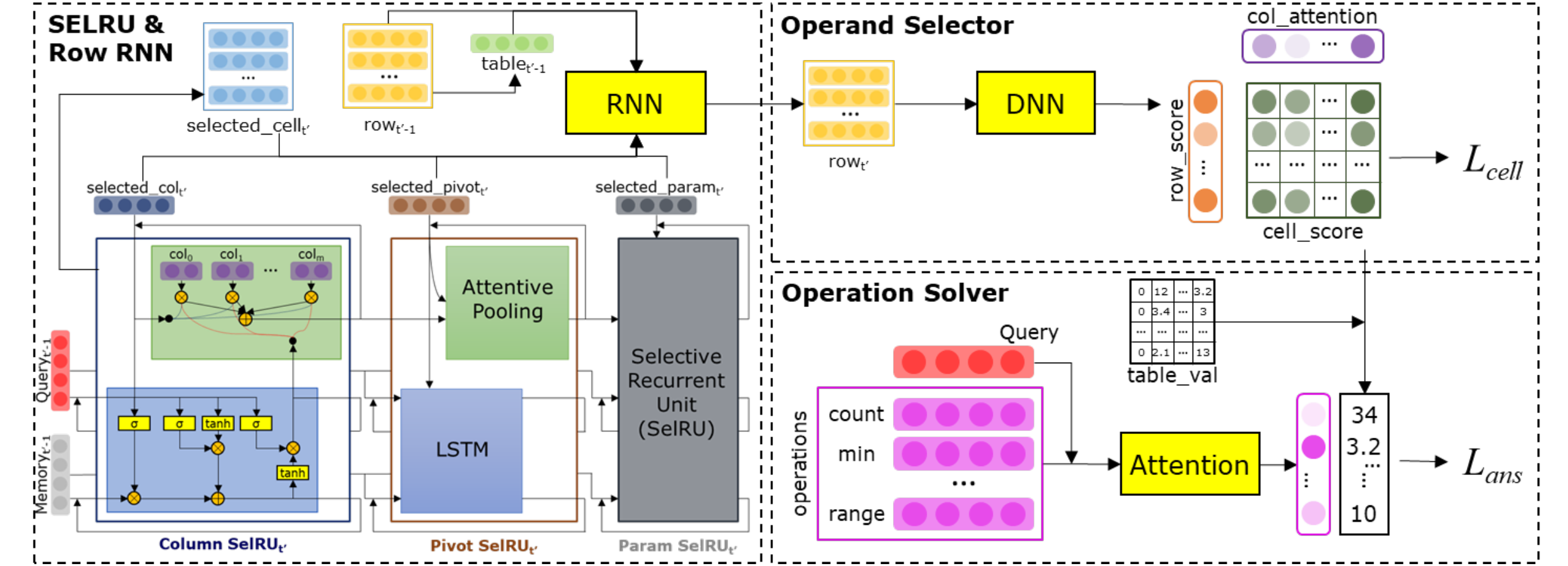}
    \caption{Full architecture of \textsc{Neural Operator} at the final timestep $t'$ of the SelRU and Row RNN.}
    \label{fig:neop}
\end{figure*}

Second, selected column vector $\widetilde{f}_{t}$ is obtained by attentively pooling the column vectors $f_k$, and selected pivot vector $\widetilde{v}_{t}$ and selected param vector $\widetilde{r}_{t}$ are calculated in similar manners as we will describe later. At timestep $t$, the selected column vector $\widetilde{f}_{t}$ is calculated as follows:

\begin{equation}
  \small
  \label{eq:att1}
  e_f (f_k, \widetilde{f}_{t-1}, q_{f_{t}}) = u_f^\top tanh(W_f [f_k;  \widetilde{f}_{t-1}; q_{f_{t}}] + b_f)
\end{equation}
\begin{equation}
  \small
  \label{eq:att}
 {a_f (f_k)}_{t} = \frac{\exp(e_f (f_k, \widetilde{f}_{t-1}, q_{f_{t}}))}{\sum_{k'} \exp(e_f (f_{k'}, \widetilde{f}_{t-1}, q_{f_{t}}))}
\end{equation}
\begin{equation}
  \small
  \label{eq:att3}
  \widetilde{f}_{t} = \sum_k {a_f (f_k)}_{t} \times f_k
\end{equation}
Here ${a_f (f_k)}_{t}$ is the attention weight of the $k$th column at timestep $t$ using two contexts: the selected column vector $\widetilde{f}_{t-1}$ at timestep ${t-1}$ and the query vector $q_{f_{t}}$ containing previous selection information from LSTM layer (Eq. \ref{eq:lstm}).
This decision is cascaded to
Pivot and Param SelRUs, in calculating the selected pivot vector and param vector
 $\widetilde{v}_{t}$ and  $\widetilde{r}_{t}$ respectively.
 Each module uses the same network as column selection, only cascading the column and pivot just decided as additional contexts, as Figure~\ref{fig:neop} shows.
 
\subsection{Row RNN}
\label{sec:rowrnn}
Once we obtain the selected column, pivot, and parameter vectors $\widetilde{f}_t$, $\widetilde{v}_t$, and $\widetilde{p}_t$ at timestep $t$, we use these vectors as the conditions to select the correct rows. To do this selection process, we use a simple RNN with multiple contexts called Row RNN to represent $n$ row vectors at timestep $t$, i.e. $r_{j_t}$. The row vectors $r_{j_t}$ represent the information whether to select the row or not. For each row $j$, Row RNN accepts six inputs: (1-3) the current selected column, pivot, and parameter vectors $\widetilde{f}_t$, $\widetilde{v}_t$, and $\widetilde{p}_t$, (4) the current selected cell vector at row $r_j$, i.e. $\widetilde{c}_{j_t}$, (5) the previous row vector $r_{j_{t-1}}$, and (6) the previous table vector $x_{t-1}$, and returns the row vector $r_{j_t}$ as output.

The selected cell vector $\widetilde{c}_{j_t}$ contains information of the row $j$ biased selected column at timestep $t$.
This is obtained by performing a weighted sum of the cell vectors $c_{j,k}$ using the column attention scores from Column SelRU (i.e. $a(f_k)$ in Eq. \ref{eq:att}), i.e. $\widetilde{c}_{j_t} = \sum_k a(f_k) \times c_{j,k}$. The table vector $x_{t}$ contains the global information regarding result of operation. This is obtained by performing a max pool on all the row vectors $r_{j_t}$, i.e. $x_t = max\_pool (r_{1_t}, ..., r_{n_t})$, where $j = 1, ..., n$.

Finally, the row vector $r_{j_t}$ is obtained through a non-linear activation function using the six inputs above, as shown in Eq. \ref{eq:row}.
\begin{equation}
    \small
    \label{eq:row}
    r_{j_t} = tanh(W_r [\widetilde{f}_t; \widetilde{v}_t; \widetilde{p}_t; \widetilde{c}_{j_t}; r_{j_{t-1}}; x_{t-1}] + b_r)
\end{equation}

\subsection{Operand Selector}

\label{sec:opsel}

After the final timestep $t'$ of Row RNN, we obtain the final row vectors $r_{j_{t'}}$. The operand selector translates these vectors into $n$ scalar row scores ${p(r)}_{t'}$ with values corresponding to the probability of selecting a row. This is obtained by transforming the concatenated version of the row vectors, using a sigmoid activation function, as shown in Eq. \ref{eq:row_score}.
\begin{equation}
    \small
    \label{eq:row_score}
    {p(r)}_{t'} = \sigma(W_a [r_{1_{t'}}; ...; r_{n_{t'}}] + b)
\end{equation}
Moreover, the operand selector uses the product of both the row scores ${p(r_j)}_{t'}$ and the column attention scores from Column SelRU ${a_f (f_k)}_{t'}$ (Eq. \ref{eq:att}) to calculate the cell scores $C(j,k)$, which represents the probability of selecting the operands used when calculating the final answer, as shown in Eq. \ref{eq:cell_score}. 
\begin{equation}
    \small
    \label{eq:cell_score}
    C(j,k) = {p(r_j)}_{t'} \times {a_f (f_k)}_{t'}
\end{equation}
At test time, we filter the cells using a threshold $\gamma$, where cells with scores $C(j,k) > \gamma$ are the selected operands.

\subsection{Operation Solver}
\label{sec:opsolve}

Finally, we use the cell scores $C(j,k)$ to solve all the operations available in the model. In the experiments, we use the seven operations\footnote{We use the operations: argmax, argmin, average, count, print, sum and range}, but our model can easily be extended to use other operations.
An operation consists of an embedding $o$ and a function $op_o$ that returns an aggregated value $y_o$ as output. 
The operation solver accepts several operations, compares them to the query $q$, selects the appropriate operation, and solves the final answer using the operation function.

Operation selection is done by computing the relative probabilities of the operations given the query and selecting the operation with the highest probability. The relative probabilities $a_o(o)$ are obtained using attention mechanism \citep{bahdanau2014neural}, as shown in Eq. \ref{eq:opsel}.
\begin{align}
    \small
    \label{eq:opsel}
    e_o (o, q) &= u_o^\top tanh(W_o [o; q] + b_o) \nonumber \\
    a_o (o) &= \frac{\exp(e_o (o, q))}{\sum_{o'} \exp(e_o (o', q))}
\end{align}

\begin{wraptable}{r}{0.5\textwidth}
    \footnotesize
    \centering
    \begin{tabular}{|l|l|}
        \hline
        Function & Definition \\ \hline
        $all(\cdot)$ & $-\infty$ \\ \hline
        $min(\cdot)$ & $max(C(j,k) * T'[j,k])$ \\ \hline
        $max(\cdot)$ & $max(C(j,k) * T[j,k])$ \\ \hline
        $count(\cdot)$ & $\sum_{j,k} C(j,k)$ \\ \hline
        $sum(\cdot)$ & $\sum_{j,k} C(j,k) * T[j,k]$\\ \hline
        $mean(\cdot)$ & $sum(\cdot) / count(\cdot)$\\ \hline
        $range(\cdot)$ & $max(\cdot) - min(\cdot)$\\ \hline
    \end{tabular}
    \caption{Operation functions and definitions. The dot $\cdot$ corresponds to the cell values $T[j,k]$ and the cell scores $C(j,k)$ parameters.}
    \label{tab:opfxn}
\end{wraptable}

At training time, to ensure that the procedure of predicting answer is differentiable, we adopt soft approximations for each of the operations using the operation functions $op_o$ defined in Table \ref{tab:opfxn}.
All functions accept as inputs all the cell scores $C(j,k)$ and the original cell values $T[j,k]$. Our model predicts the final answer $y'$, as shown in Eq. \ref{eq:opresult}.
There are three main challenges here and we solve them through the following: (1) to handle non-numerical cell values in $T[j,k]$, we set them to zero; (2) since \texttt{all} operation is a non-numerical operation, we set its function to $-\infty$; and (3) we define reversed cell values as $T'[j,k] = max(T[j,k]) - T[j,k] + \epsilon$ to be used when the operation requires an extremity opposite to the extremity of the cell scores $C(j,k)$, such as the \texttt{min} operation. The bias parameter $\epsilon$ can be any positive number; we set ours to one. At test time, we use the original functions of the operations to get the final answer. 
\begin{align}
    \small
    \label{eq:opresult}
    y' &= \sum_{o'} a_o(o') \times op_{o'}(\cdot)
\end{align}
\subsection{Training objective}

To train the model, we use two loss functions: $L_{cell}$ for selecting the operands and $L_{ans}$ for solving the final answer. We use the log-loss function over all the cells to calculate $L_{cell}$, as shown in Eq. \ref{eq:op_loss}, where $I[j,k]$ is the labeled operand information. For $L_{ans}$, we use L2 loss between the predicted and the actual answer $y$. We notice that $L_{ans}$, as it is, is significantly larger than $L_{cell}$. Thus, we use natural logarithm on top of L2 loss to balance the weight intensity of two loss, as shown in Eq. \ref{eq:ans_loss}. The final loss is the sum of both loss functions, as shown in Eq. \ref{eq:final_loss}.
\begin{align}
    \small
    \label{eq:op_loss}
    L_{cell} &= \sum_{j,k}\Big(I[j,k] \log{C(j,k}) + 
    (1-I[j,k]) \log{\big(1-C(j,k)\big)\Big)} \\
    \label{eq:ans_loss}
    L_{ans} &= \log{\Big(\sum_i{(y'_i - y_i)}^2\Big)} \\
    \label{eq:final_loss}
    L &= L_{cell} + L_{ans}
\end{align}
\section{Experiments}
\label{sec:experiments}
\subsection{Experimental settings}

We use a 600-dimension embedding matrix for our word embedding: 300 dimensions are used for the Xavier-initialized embedding and the other 300 dimensions are used for the binary encoding. For the GRU query encoder, we set the state size to 300 and the timestep to 4, following \citep{neelakantan2016learning}. We use dropout \citep{srivastava2014dropout} on all non-linear connections with a dropout rate of 0.2. We set the batch size to 50. Training is done via stochastic gradient descent with the Adadelta update rule \citep{zeiler2012adadelta}, with $l_2$ constraint \citep{hinton2012improving} of 3. We tune the operand threshold $\gamma$ using a separate development set and find that anywhere between 0.4 and 0.6 provides similar performance. We set $\gamma = 0.5$ during test phase.

\subsection{\textsc{MLB} dataset}

\paragraph{Competing models and evaluation}

\begin{table}[t]
    \footnotesize
    \centering
    \begin{tabular}{|c|ccc|c|}
    \hline

& $SoftOpP$ & $SoftOpR$ & $HardOpA$ & $FinalAcc$ \\
    \hline
    \textsc{Sempre} & 51.3 & 60.1 & 58.4 & 64.4 \\
    \textsc{NePr} & 22.8 & 29.6 & 22.9 & 32.4 \\
    \textsc{NeEn} & 62.7 & 6.4 & 8.3 & 13.3 \\ \hline
    \textbf{\textsc{NeOp} (ours)} & \textbf{92.9} & \textbf{94.1} & \textbf{72.3} & \textbf{80.3} \\
    \hline
    \end{tabular}%

  \caption{Results of competing models on \textsc{MLB} dataset. Top scores are \textbf{bold-faced}.}
  \label{tab:results_mlb}%
\end{table}%

We compare our models with three competing TableQA models: one semantic parsing-based model, \textsc{Sempre} \citep{pasupat2015compositional}, and two neural-based models, Neural Enquirer \textsc{NeEn} \citep{yin2016neural} and Neural Programmer \textsc{NePr} \citep{neelakantan2016learning}. We use the \textsc{MLB} dataset and it consists of 16k train set, 10k development set and 10k test set.
We evaluate the models using four metrics: (1-2) soft operand precision and range when selecting a single cell as an operand, i.e. $SoftOpP = \frac{\#CorrectSelected}{\#TotalSelected}$ and $SoftOpR = \frac{\#CorrectSelected}{\#TotalOperands}$; (3) hard operand accuracy is the percentage of correct prediction of operand \textit{sets}, i.e. $HardOpA = \frac{\#CorrectSets}{\#TotalSets}$; and (4) final answer accuracy is the percentage of correct final answers, i.e. $FinalAcc = \frac{\#CorrectAnswers}{\#TotalAnswers}$.

\paragraph{Results}

We report the results in Table \ref{tab:results_mlb}. 
\textsc{NeOp} performs the best on all metrics, outperforming the second best model, \textsc{Sempre}, by at least 26\%. 
\textsc{NeEn} performs the worst in the final accuracy metric, even though it obtains good soft operand precision. This is because \textsc{NeEn} is not able to answer questions that need multiple cells.

Another interesting observation is that \textsc{NePr} achieves a final accuracy much higher than the hard operand accuracy.
This means that even though the models selected the wrong cells, they are magically able to get the answer correctly, a similar anomaly example in Figure \ref{fig:example}. This hurts the interpretability of the models and the robustness of their performance. We study these issues more closely in Section 5.5.
On the other hand, the final accuracy and the hard operand accuracy of \textsc{NeOp} have minimal difference, thus making the results reliable.

\vspace{4cm}

\subsection{\textsc{WikiOps} dataset}

\begin{wraptable}{r}{0.5\textwidth}
    \footnotesize 
    \centering
    {
    \centering
    \begin{tabular}{|c|c|c|}
        \hline
        Dataset & Model & $Final Acc$ \\ \hline
        \multirow{3}[1]{*} {\textsc{WikiOps}} & \textbf{NeOp (ours)} & \textbf{59.5} \\
        \cline{2-3}
         & \textsc{NePr} & 52.0 \\ \cline{2-3}
         & \textsc{NeEn} & 3.4 \\ \hline
         
         \multirow{3}[1]{*} {\textsc{WikiSQL}} & \textsc{Seq2SQL} & \textbf{59.4}\footnotemark\\
        \cline{2-3}
         & \textsc{AugPtr} & 53.3 \\ \cline{2-3}
         & \textsc{Seq2Seq} & 35.9 \\ \hline
        
    \end{tabular}
    }
    \caption{Results of models on \textsc{WikiOps} and \textsc{WikiSQL}. Top scores are \textbf{bold-faced}}
    \label{tab:results_wikiops}
\end{wraptable}
\footnotetext{ \textsc{Seq2SQL} is being outdated with newer results, but our intention of reporting this number is not to compete, considering their unfair advantages of using SQL statements as more informative supervisions. 
         Considering its major limitations as explained in Section \ref{sec:intro}, we report this rather as an oracle result.}

\paragraph{Competing models and evaluation}

We compare our model with five competing TableQA models:
two neural-based models\footnote{We are not able to run \textsc{Sempre} due to about 7x higher training complexity than neural-based model, which makes the training infeasible for large sets.}, Neural Enquirer \textsc{NeEn} \citep{yin2016neural} and Neural Programmer \textsc{NePr} \citep{neelakantan2016learning}, 
and three models from the \textsc{WikiSQL} dataset:
Sequence-to-Tree with Attention \textsc{Seq2Tree} \citep{dong2016language}, the Augmented Pointer Network \textsc{AugPtr} \citep{zhongSeq2SQL2017} and the Sequence-to-SQL \textsc{Seq2SQL} \citep{zhongSeq2SQL2017}.
We use the same train/development/test split of the original \textsc{WikiSQL} dataset.
Since some models do not have SQL statement or operand information supervision, we only compare the models using the final answer accuracy, i.e. $FinalAcc = \frac{\#CorrectAnswers}{\#TotalAnswers}$.

\paragraph{Results}

The results are reported in Table \ref{tab:results_wikiops}. Using the \textsc{WikiOps} dataset, \textsc{NeOp} outperforms all other models. The \textsc{NePr} model performs relatively good compared to its performance in the \textsc{MLB} dataset. We argue this is because of the dataset distribution of \textsc{WikiOps} dataset that is skewed towards a lot of \texttt{all} operations.
Nevertheless, our model still performs better than \textsc{NeOp} by 7.5\%. Finally, \textsc{NeEn} performs the worst again in this dataset, since this dataset also contains queries that need multiple cells.

When compared with models using the \textsc{WikiSQL} dataset, \textsc{NeOp} outperforms two baseline models \textsc{Seq2Tree} and \textsc{AugPtr}, and performs comparably with the \textsc{Seq2SQL} model, despite unfair disadvantage of competitors being supervised by SQL statements that are more informative. 

\subsection{Analysis on operand loss}

We argue that one of the reasons why \textsc{NeOp} performs well is that we use the operand information for solving the queries.
We perform further analysis on the use of operand information by adding/removing the operand loss on two models: \textsc{NeEn} and \textsc{NeOp}. We add the operand loss to the objective function of \textsc{NeEn} to create an improved version of the model. Furthermore, we remove the operand loss to the objective function of \textsc{NeOp} and obtain a reduced version of the model. We report the results in Table \ref{tab:op_results} on the \textsc{MLB} dataset. Both models enjoy increases in performance on all metrics. This is especially noticeable in the hard operand accuracy, where \textsc{NeEn} and \textsc{NeOp} receive a 139.8\% and 107.8\% increase, respectively, and in the final accuracy, where \textsc{NeEn} and \textsc{NeOp} receive a 67.7\% and 47.1\% increase, respectively.

\begin{table}[h]
  \footnotesize
  \centering
    \begin{tabular}{|c|cc|cc|}
    \hline
    \multirow{2}[1]{*}{~~~Model~~~} & \multicolumn{2}{c|}{$SoftOpP$} & \multicolumn{2}{c|}{$SoftOpR$} \\
\cline{2-5}      & w/o $L_{cell}$ & w/ $L_{cell}$ & w/o $L_{cell}$ & w/ $L_{cell}$ \\
    \hline
    \textsc{NeEn} & 62.7 & \textbf{94.5 (+50.7\%)~~} & 6.4 & \textbf{9.6 (+50.0\%)} \\
    \textsc{NeOp} & 83.4 & \textbf{92.9 (+11.4\%)~~} & 80.0 & \textbf{94.1 (+17.6\%)} \\
    \hline
    \end{tabular}%
    \\
    \centering
    \begin{tabular}{|c|cc|cc|}
    \hline
    \multirow{2}[1]{*}{~~~Model~~~} & \multicolumn{2}{c|}{$HardOpA$} & \multicolumn{2}{c|}{$FinalAcc$} \\
\cline{2-5}      & w/o $L_{cell}$ & w/ $L_{cell}$ & w/o $L_{cell}$ & w/ $L_{cell}$ \\
    \hline
    \textsc{NeEn} & 8.3 & \textbf{19.9 (+139.8\%)} & 13.3 & \textbf{22.3 (+67.7\%)} \\
    \textsc{NeOp} & 34.8 & \textbf{72.3 (+107.8\%)} & 54.6 & \textbf{80.3 (+47.1\%)} \\
    \hline
    \end{tabular}
  \caption{Increase in performance of \textsc{NeEn} and \textsc{NeOp} when $L_{cell}$ is introduced.}
  \label{tab:op_results}%
\end{table}

\subsection{Robustness to adversarial examples}
\label{sec:adversarial}

From the results in Table \ref{tab:results_mlb}, we could observe that the models predict the correct answer with inaccurate cell selection. 
To examine how such ``lucky guesses'' hurt the robustness and interpretability of the models, we
 manually created adversarial examples and evaluate the models with such examples. Specifically,
 we study two classes of lucky guess scenarios, with data and operation perturbations:
 \begin{itemize}
 \item Value Perturbation (V-P) (perturbing 12\% of test set):
From the results in Table \ref{tab:results_mlb},
we find many answers coincide as $0$, such that many lucky guesses generated the right answer, merely based on frequency. We change the values that do not affect the answers to reduce its frequency.
 \item Operation Perturbation (O-P) (perturbing 70\% of test set):
 Some operation, such as count, can be answered correctly, regardless of the columns selected. 
We change the operation
(e.g. replace \texttt{argmax} to \texttt{sum}) and answer accordingly.
\end{itemize}

\begin{table}[t]
  \small
  \centering
    \begin{tabular}{|c|c|cc|cc|}
    \hline
     Model & $Final Acc$ & V-P & O-P \\
    \hline
    \textsc{SEMPRE} & 64.4 & \textbf{62.6 (-2.80\%)~~} & \textbf{64.2 (-0.36\%)~~} \\
    \textsc{NePR} & 32.4 & \textbf{31.2 (-3.70\%)~~} & \textbf{32.0 (-1.23\%)~~}  \\
    \textsc{NeOp} & 80.3 & \textbf{80.1 (-0.25\%)~~} & \textbf{80.1 (-0.25\%)~~}  \\
    \hline
    \end{tabular}%
    \caption{Results of models on adversarial examples \textsc{MLB} dataset.}
  \label{tab:ad_results}%
\end{table}

\paragraph{Results}

Using the perturbed dataset,
we compare our model \textsc{NeOp} with semantic parsing-based model, \textsc{Sempre} \citep{pasupat2015compositional} and neural-based models, \textsc{NePr} \citep{neelakantan2016learning}.
For V-P, \textsc{NeOp} stays robust, only with a slight degradation (0.25\%) from its original accuracy.
However, \textsc{Sempre} and \textsc{NeOp} show more significant decreases of 2.80\% and 3.70\% from their original accuracy, respectively. 
For O-P, \textsc{NePr} is still the most sensitive to adversarial examples with the decrease of 1.23\%. \textsc{NeOp} and \textsc{Sempre} 
responds better to adversarial operation changes.

\subsection{Model interpretability}
\label{sec:interpret}

Through the Column, Pivot, Param SelRUs, and the Operation Solver of \textsc{NeOp}, it is easy to interpret the model in a step-by-step manner.
First, at each timestep, we obtain the attention weights of the Attentive Pooling component of each SelRU (e.g. Eq. \ref{eq:att} of Column SelRU). The one with the highest weight is the selected field/word.
Second, after the final timestep, we obtain the attention weights of the operation solver (i.e. $a_o(o)$ in Eq. \ref{eq:opsel}). The one with the highest weight is the selected operation.

\begin{figure}
    \centering
    \includegraphics[width=0.5\textwidth]{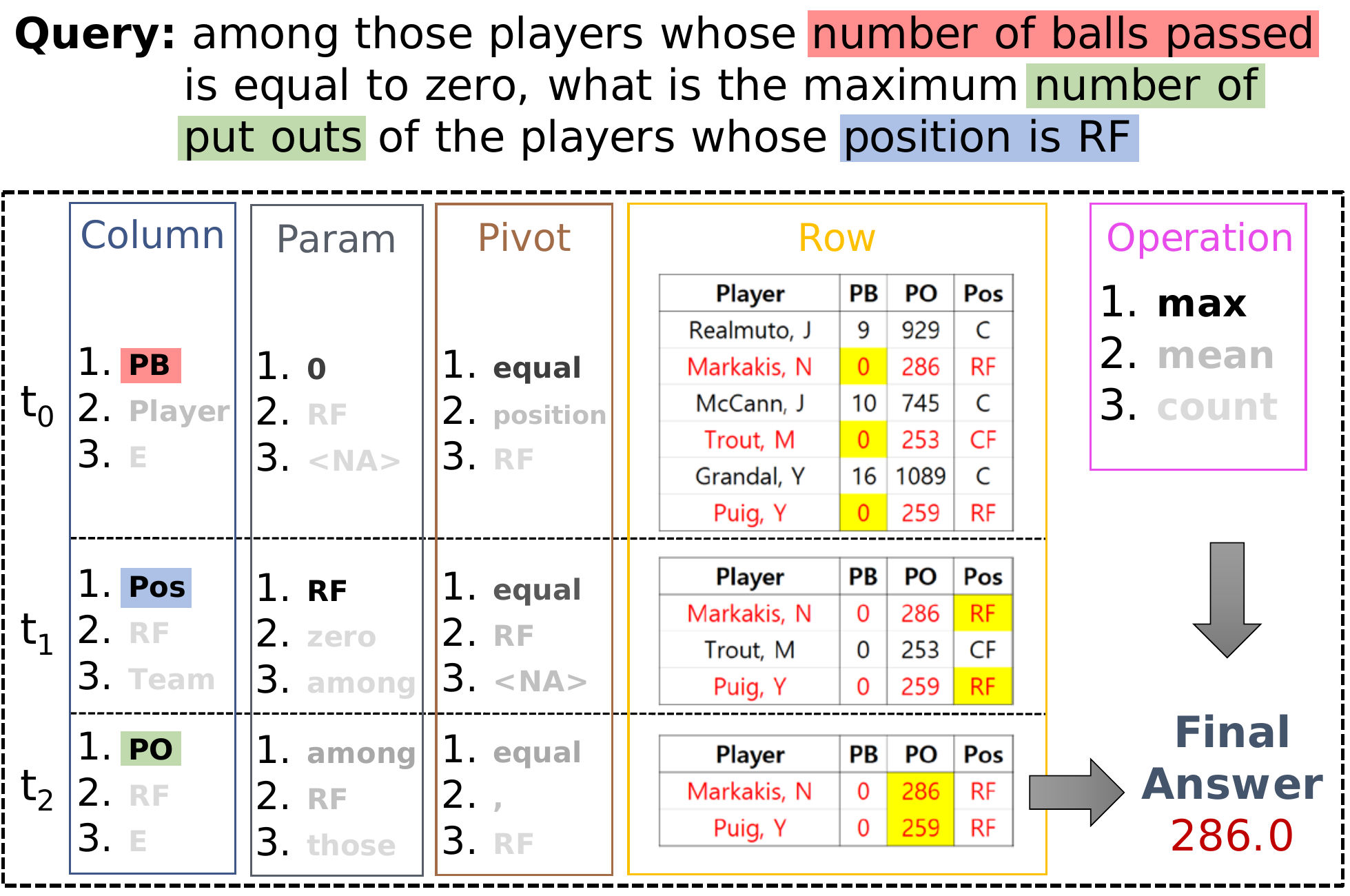}
    \caption{An example query ran through \textsc{NeOp}. The columns, pivots, parameters, and the operation are selected using the weights from our model. We show three instances with the highest weights at each timestep. The last timestep is omitted.
    The rows in the right side are manually interpreted.
    The logical form of the natural language query shown is \texttt{max(PO | PB=0, Pos='RF')}. 
    Abbreviation glossary: PO - put outs, PB - passed balls, Pos - position.}
    \label{fig:interpret}
\end{figure}

In Figure \ref{fig:interpret}, we show an example query answered by our model.
At the first timestep, the selected column \texttt{PB}, the selected pivot is the word ``\textit{equal}'', and the selected parameter is the word ``\textit{0}''. These elements correspond to the first condition of the query, i.e. \texttt{PB=0}. At the second timestep, a similar step-by-step selection of column, pivot, and parameter is carried out to create the second condition of the query, i.e. \texttt{Pos='RF'}. Finally, at the last timestep, although the words ``\textit{equal}'' and ``\textit{among}'' are selected as pivot and parameter, the weights are not high enough. Hence, this signals the end of the iteration of the SelRUs.
A possible interpretation is shown in the yellow box at the right of the same figure, where selected rows are colored red and conditioned cells are highlighted in yellow.
Finally, the operation solver selects \texttt{max} operation, thus the final answer is \texttt{max(286, 259) = 286}.

\section{Conclusion}

In this paper, we propose the usage of operand information to improve the performance of TableQA models. To do this, we create two new datasets called \textsc{MLB} dataset and \textsc{WikiOps} dataset, both of which improves on previous datasets in terms of attention supervision. Furthermore, we develop \textsc{Neural Operator}, a neural-based TableQA model which has improved interpretability thanks to the its layer-wise architecture and the use of operand information. We test our hypothesis on both proposed datasets, and our model significantly outperforms previous models. Finally, we show multiple analysis on how the operand information increases the performance, reliability, and interpretability of models.

\bibliographystyle{named}
\bibliography{acml18}

\begin{thebibliography}{32}
\providecommand{\natexlab}[1]{#1}
\providecommand{\url}[1]{\texttt{#1}}
\expandafter\ifx\csname urlstyle\endcsname\relax
  \providecommand{\doi}[1]{doi: #1}\else
  \providecommand{\doi}{doi: \begingroup \urlstyle{rm}\Url}\fi

\bibitem[Amplayo et~al.(2018{\natexlab{a}})Amplayo, Kim, Sung, and won
  Hwang]{Amplayo2018ColdStartAU}
Reinald~Kim Amplayo, Jihyeok Kim, Sua Sung, and Seung won Hwang.
\newblock Cold-start aware user and product attention for sentiment
  classification.
\newblock In \emph{ACL}, 2018{\natexlab{a}}.

\bibitem[Amplayo et~al.(2018{\natexlab{b}})Amplayo, Lee, Yeo, and won
  Hwang]{Amplayo2018TranslationsAA}
Reinald~Kim Amplayo, Kyungjae Lee, Jinyeong Yeo, and Seung won Hwang.
\newblock Translations as additional contexts for sentence classification.
\newblock In \emph{IJCAI}, 2018{\natexlab{b}}.

\bibitem[Amplayo et~al.(2018{\natexlab{c}})Amplayo, Lim, and won
  Hwang]{amplayo2017entity}
Reinald~Kim Amplayo, Seonjae Lim, and Seung won Hwang.
\newblock Entity commonsense representation for neural abstractive
  summarization.
\newblock In \emph{NAACL}, 2018{\natexlab{c}}.

\bibitem[Bahdanau et~al.(2014)Bahdanau, Cho, and Bengio]{bahdanau2014neural}
Dzmitry Bahdanau, Kyunghyun Cho, and Yoshua Bengio.
\newblock Neural machine translation by jointly learning to align and
  translate.
\newblock \emph{arXiv preprint arXiv:1409.0473}, 2014.

\bibitem[Berant et~al.(2013)Berant, Chou, Frostig, and
  Liang]{berant2013semantic}
Jonathan Berant, Andrew Chou, Roy Frostig, and Percy Liang.
\newblock Semantic parsing on freebase from question-answer pairs.
\newblock In \emph{Proceedings of the 2013 Conference on Empirical Methods in
  Natural Language Processing}, pages 1533--1544, 2013.

\bibitem[Chung et~al.(2014)Chung, Gulcehre, Cho, and
  Bengio]{chung2014empirical}
Junyoung Chung, Caglar Gulcehre, KyungHyun Cho, and Yoshua Bengio.
\newblock Empirical evaluation of gated recurrent neural networks on sequence
  modeling.
\newblock \emph{arXiv preprint arXiv:1412.3555}, 2014.

\bibitem[Cui et~al.(2017)Cui, Xiao, Wang, Song, Hwang, and Wang]{cui2017kbqa}
Wanyun Cui, Yanghua Xiao, Haixun Wang, Yangqiu Song, Seung-won Hwang, and Wei
  Wang.
\newblock Kbqa: learning question answering over qa corpora and knowledge
  bases.
\newblock \emph{Proceedings of the VLDB Endowment}, 10\penalty0 (5):\penalty0
  565--576, 2017.

\bibitem[Das et~al.(2017)Das, Agrawal, Zitnick, Parikh, and
  Batra]{das2017human}
Abhishek Das, Harsh Agrawal, Larry Zitnick, Devi Parikh, and Dhruv Batra.
\newblock Human attention in visual question answering: Do humans and deep
  networks look at the same regions?
\newblock \emph{Computer Vision and Image Understanding}, 163:\penalty0
  90--100, 2017.

\bibitem[Dong and Lapata(2016)]{dong2016language}
Li~Dong and Mirella Lapata.
\newblock Language to logical form with neural attention.
\newblock \emph{arXiv preprint arXiv:1601.01280}, 2016.

\bibitem[Glorot and Bengio(2010)]{glorot2010understanding}
Xavier Glorot and Yoshua Bengio.
\newblock Understanding the difficulty of training deep feedforward neural
  networks.
\newblock In \emph{Proceedings of the Thirteenth International Conference on
  Artificial Intelligence and Statistics}, pages 249--256, 2010.

\bibitem[Goldman et~al.(2017)Goldman, Latcinnik, Naveh, Globerson, and
  Berant]{goldman2017weakly}
Omer Goldman, Veronica Latcinnik, Udi Naveh, Amir Globerson, and Jonathan
  Berant.
\newblock Weakly-supervised semantic parsing with abstract examples.
\newblock \emph{arXiv preprint arXiv:1711.05240}, 2017.

\bibitem[Hinton et~al.(2012)Hinton, Srivastava, Krizhevsky, Sutskever, and
  Salakhutdinov]{hinton2012improving}
Geoffrey~E Hinton, Nitish Srivastava, Alex Krizhevsky, Ilya Sutskever, and
  Ruslan~R Salakhutdinov.
\newblock Improving neural networks by preventing co-adaptation of feature
  detectors.
\newblock \emph{arXiv preprint arXiv:1207.0580}, 2012.

\bibitem[Jia and Liang(2017)]{jia2017adversarial}
Robin Jia and Percy Liang.
\newblock Adversarial examples for evaluating reading comprehension systems.
\newblock \emph{arXiv preprint arXiv:1707.07328}, 2017.

\bibitem[Krishnamurthy and Kollar(2013)]{krishnamurthy2013jointly}
Jayant Krishnamurthy and Thomas Kollar.
\newblock Jointly learning to parse and perceive: Connecting natural language
  to the physical world.
\newblock \emph{Transactions of the Association for Computational Linguistics},
  1:\penalty0 193--206, 2013.

\bibitem[Kwiatkowski et~al.(2013)Kwiatkowski, Choi, Artzi, and
  Zettlemoyer]{kwiatkowski2013scaling}
Tom Kwiatkowski, Eunsol Choi, Yoav Artzi, and Luke Zettlemoyer.
\newblock Scaling semantic parsers with on-the-fly ontology matching.
\newblock In \emph{2013 Conference on Empirical Methods in Natural Language
  Processing, EMNLP 2013}. Association for Computational Linguistics (ACL),
  2013.

\bibitem[Liu et~al.(2017{\natexlab{a}})Liu, Mao, Sha, and
  Yuille]{liu2017attention}
Chenxi Liu, Junhua Mao, Fei Sha, and Alan~L Yuille.
\newblock Attention correctness in neural image captioning.
\newblock In \emph{AAAI}, pages 4176--4182, 2017{\natexlab{a}}.

\bibitem[Liu et~al.(2017{\natexlab{b}})Liu, Shen, Duh, and
  Gao]{liu2017stochastic}
Xiaodong Liu, Yelong Shen, Kevin Duh, and Jianfeng Gao.
\newblock Stochastic answer networks for machine reading comprehension.
\newblock \emph{arXiv preprint arXiv:1712.03556}, 2017{\natexlab{b}}.

\bibitem[Mi et~al.(2016)Mi, Wang, and Ittycheriah]{mi2016supervised}
Haitao Mi, Zhiguo Wang, and Abe Ittycheriah.
\newblock Supervised attentions for neural machine translation.
\newblock In \emph{Proceedings of the 2016 Conference on Empirical Methods in
  Natural Language Processing}, pages 2283--2288, 2016.

\bibitem[Min et~al.(2017)Min, Seo, and Hajishirzi]{min2017question}
Sewon Min, Minjoon Seo, and Hannaneh Hajishirzi.
\newblock Question answering through transfer learning from large fine-grained
  supervision data.
\newblock \emph{arXiv preprint arXiv:1702.02171}, 2017.

\bibitem[Neelakantan et~al.(2015)Neelakantan, Le, and
  Sutskever]{neelakantan2015neural}
Arvind Neelakantan, Quoc~V Le, and Ilya Sutskever.
\newblock Neural programmer: Inducing latent programs with gradient descent.
\newblock \emph{arXiv preprint arXiv:1511.04834}, 2015.

\bibitem[Neelakantan et~al.(2016)Neelakantan, Le, Abadi, McCallum, and
  Amodei]{neelakantan2016learning}
Arvind Neelakantan, Quoc~V Le, Martin Abadi, Andrew McCallum, and Dario Amodei.
\newblock Learning a natural language interface with neural programmer.
\newblock \emph{arXiv preprint arXiv:1611.08945}, 2016.

\bibitem[Pasupat and Liang(2015)]{pasupat2015compositional}
Panupong Pasupat and Percy Liang.
\newblock Compositional semantic parsing on semi-structured tables.
\newblock \emph{arXiv preprint arXiv:1508.00305}, 2015.

\bibitem[Rajpurkar et~al.(2016)Rajpurkar, Zhang, Lopyrev, and
  Liang]{rajpurkar2016squad}
Pranav Rajpurkar, Jian Zhang, Konstantin Lopyrev, and Percy Liang.
\newblock Squad: 100,000+ questions for machine comprehension of text.
\newblock \emph{arXiv preprint arXiv:1606.05250}, 2016.

\bibitem[Srivastava et~al.(2014)Srivastava, Hinton, Krizhevsky, Sutskever, and
  Salakhutdinov]{srivastava2014dropout}
Nitish Srivastava, Geoffrey Hinton, Alex Krizhevsky, Ilya Sutskever, and Ruslan
  Salakhutdinov.
\newblock Dropout: A simple way to prevent neural networks from overfitting.
\newblock \emph{The Journal of Machine Learning Research}, 15\penalty0
  (1):\penalty0 1929--1958, 2014.

\bibitem[Wang et~al.(2015)Wang, Berant, Liang, et~al.]{wang2015building}
Yushi Wang, Jonathan Berant, Percy Liang, et~al.
\newblock Building a semantic parser overnight.
\newblock In \emph{ACL (1)}, pages 1332--1342, 2015.

\bibitem[Yang et~al.(2015)Yang, Yih, and Meek]{yang2015wikiqa}
Yi~Yang, Wen-tau Yih, and Christopher Meek.
\newblock Wikiqa: A challenge dataset for open-domain question answering.
\newblock In \emph{EMNLP}, pages 2013--2018, 2015.

\bibitem[Yin et~al.(2016)Yin, Lu, Li, and Kao]{yin2016neural}
Pengcheng Yin, Zhengdong Lu, Hang Li, and Ben Kao.
\newblock Neural enquirer: learning to query tables in natural language.
\newblock In \emph{Proceedings of the Twenty-Fifth International Joint
  Conference on Artificial Intelligence}, pages 2308--2314. AAAI Press, 2016.

\bibitem[Yu et~al.(2017)Yu, Choi, Kim, Yoo, Lee, and Kim]{yu2017supervising}
Youngjae Yu, Jongwook Choi, Yeonhwa Kim, Kyung Yoo, Sang-Hun Lee, and Gunhee
  Kim.
\newblock Supervising neural attention models for video captioning by human
  gaze data.
\newblock In \emph{IEEE Conference on Computer Vision and Pattern Recognition
  (CVPR 2017). Honolulu, Hawaii}, pages 2680--29, 2017.

\bibitem[Zeiler(2012)]{zeiler2012adadelta}
Matthew~D Zeiler.
\newblock Adadelta: an adaptive learning rate method.
\newblock \emph{arXiv preprint arXiv:1212.5701}, 2012.

\bibitem[Zettlemoyer and Collins(2005)]{zettlemoyer2005learning}
Luke~S Zettlemoyer and Michael Collins.
\newblock Learning to map sentences to logical form: Structured classification
  with probabilistic categorial grammars.
\newblock In \emph{In Proceedings of the 21st Conference on Uncertainty in AI}.
  Citeseer, 2005.

\bibitem[Zhao and Mao(2017)]{zhao2017topic}
Rui Zhao and Kezhi Mao.
\newblock Topic-aware deep compositional models for sentence classification.
\newblock \emph{IEEE/ACM Transactions on Audio, Speech, and Language
  Processing}, 25\penalty0 (2):\penalty0 248--260, 2017.

\bibitem[Zhong et~al.(2017)Zhong, Xiong, and Socher]{zhongSeq2SQL2017}
Victor Zhong, Caiming Xiong, and Richard Socher.
\newblock Seq2sql: Generating structured queries from natural language using
  reinforcement learning.
\newblock \emph{CoRR}, abs/1709.00103, 2017.

\end{thebibliography}
\end{document}